# Inter-Scanner Harmonization of High Angular Resolution DW-MRI using Null Space Deep Learning

Vishwesh Nath[1], Prasanna Parvathaneni[1], Colin B. Hansen[1], Allison E. Hainline[3], Camilo Bermudez[2], Samuel Remedios[4], Justin A. Blaber[1], Kurt G. Schilling[2], Ilwoo Lyu[1], Vaibhav Janve[2], Yurui Gao[2], Iwona Stepniewska[5], Baxter P. Rogers[6], Allen T. Newton[6], L. Taylor Davis[7], Jeff Luci[8], Adam W. Anderson[2] and Bennett A. Landman[1,2]

[1] EECS, Vanderbilt University, Nashville TN 37203, USA
[2] BME, Vanderbilt University, Nashville TN 37203, USA
[3] Biostatistics, Vanderbilt University, Nashville TN 37203, USA
[4] Computer Science, Middle Tennessee State University, Murfressboro TN 37132, USA
[5] Psychology, Vanderbilt University, Nashville TN 37203, USA
[6] VUIIS, Vanderbilt University, Nashville, TN 37232, USA
[7] VUMC, Vanderbilt University, Nashville, TN, 37203 USA
[8] BME, University of Texas at Austin, Austin, TX 78712

**Abstract.** Diffusion-weighted magnetic resonance imaging (DW-MRI) allows for non-invasive imaging of the local fiber architecture of the human brain at a millimetric scale. Multiple classical approaches have been proposed to detect both single (e.g., tensors) and multiple (e.g., constrained spherical deconvolution, CSD) fiber population orientations per voxel. However, existing techniques generally exhibit low reproducibility across MRI scanners. Herein, we propose a data-driven technique using a neural network design which exploits two categories of data. First, training data were acquired on three squirrel monkey brains using ex-vivo DW-MRI and histology of the brain. Second, repeated scans of human subjects were acquired on two different scanners to augment the learning of the network proposed. To use these data, we propose a new network architecture, the null space deep network (NSDN), to simultaneously learn on traditional observed/truth pairs (e.g., MRI-histology voxels) along with repeated observations without a known truth (e.g., scan-rescan MRI). The NSDN was tested on twenty percent of the histology voxels that were kept completely blind to the network. NSDN significantly improved absolute performance relative to histology by 3.87% over CSD and 1.42% over a recently proposed deep neural network approach. Moreover, it improved reproducibility on the paired data by 21.19% over CSD and 10.09% over a recently proposed deep approach. Finally, NSDN improved generalizability of the model to a third *in vivo* human scanner (which was not used in training) by 16.08% over CSD and 10.41% over a recently proposed deep learning approach. This work suggests that data-driven approaches for local fiber reconstruction are more reproducible, informative and precise and offers a novel, practical method for determining these models.

## 1     Introduction

Diffusion-weighted MRI (DW-MRI) provides orientation and acquisition-dependent imaging contrasts that are uniquely sensitive to the tissue microarchitecture at a millimeter scale [1]. Substantial effort has gone into modeling the relationship between observed signals and underlying biology, with a tensor model of Gaussian processes being the most commonly used model [2]. Voxel-wise models that characterize higher order spatial dependence than tensors fall under the moniker of higher angular resolution diffusion imaging (HARDI) [3]. Recently, a myriad of techniques has emerged to estimate local structure from these diffusion measures [4-7]. However, broad adoption and clinical translation of specific methods has been hindered by a lack of reproducibility [8, 9], inter-scanner stability [10, 11] , and anatomical specificity when compared to a histologically defined true microarchitecture [24]. There are known critical issues of the inter-scanner diffusion harmonization that go beyond noise effects [12-14].

Recently, it has become feasible to apply a data-driven approach to estimate tissue microarchitecture from *in vivo* diffusion weighted MRI using deep learning [25]. This approach relied on a histologically defined truth with correspondingly paired voxels with diffusion weighted magnetic resonance imaging data. Yet, no approaches to date have addressed inter-scanner variation and scan-rescan reproducibility. Moreover, traditional deep learning architecture do not specifically create models that have these necessary characteristics for clinical translation. Here, we propose a new learning architecture, the null space deep network (NSDN), to address the short comings of precision and reproducibility across scanners. Within the NSDN framework, we use inter-scanner paired *in vivo* human data to stabilize the data driven approach linking preclinical DW-MRI with histological data. Using a withheld dataset, the NSDN method is compared against a previously published fully connected network and the leading model-based approach, super resolved constrained spherical deconvolution (CSD) [4] in terms of the precision with which the model captures histologically defined truth from DW-MRI data, the reproducibility of the approach on *in vivo* human data, and the generalizability of the model to *in vivo* data acquired on an additional MRI scanner.

The remainder of this manuscript is organized as follows. Section 2 presents the acquisition and processing of all the data that has been used for the study. Section 3 presents the design and the parameters of the proposed network architecture. Section 4 presents the results. Section 5 presents the conclusion.

## 2     Data Acquisition and Processing

Three *ex-vivo* squirrel monkey brains were imaged on a Varian 9.4T scanner (Fig. 1). A total of 100 gradient volumes were acquired using a diffusion-weighted EPI sequence at a diffusivity value of 6000 s/mm$^2$, acquired at an isotropic resolution of 0.3mm. Once acquired, the tissue was sectioned and stained with fluorescent dil and imaged on a LSM710 Confocal microscope following procedures outlined in [24]. A similar procedure is outlined by [15]. The histological fiber orientation distribution (HFOD) was extracted using 3D structure tensor analysis. A multi-step registration procedure was used to determine the corresponding diffusion MRI signal. A total of 567 histological voxels were processed. 54 voxels of these were labelled as outliers qualitatively and were rejected from the analysis. A hundred random rotations were applied to the remaining voxels for both the MR signal and the HFOD to augment the data and bringing the total to 51,813 voxels [16]. A withheld set of 72 test voxels was maintained for validation. With rotations, these total to 7,272 voxels.

The *in vivo* acquisitions of the human subjects' data were acquired on three different sites, referred to as 'A', 'B' and 'C'. Three healthy human subjects were acquired with a scan each at the sites in the following manner. Subject 1: Site 'A' and Site 'B'. Subject 2: site 'A' and site 'B'. Subject 3: site 'B' and site 'C'. Structural T1 MPRAGE were acquired for all subjects at all sites. The diffusion acquisition protocol and scanner information are listed for each of the sites as follows.

Site 'A' was equipped with a 3T scanner with a 32-channel head coil. The scan was acquired at a diffusivity value of 2000 s/mm$^2$ (approximating diffusion contrast of fixed ex-vivo scan at a b-value of 6000 s/mm$^2$). 96 diffusion weighted gradient volumes were acquired with a 'b0'. Briefly the other parameters are: SENSE=2.5, partial Fourier=0.77, FOV=96x96, Slice=48, isotropic resolution: 2.5mm.

Site 'B' was equipped with a 3T scanner with a 32-channel head coil. All the parameters of the scan acquisition were as of scanner at site 'A' except for the isotropic resolution which was 1.9mmx1.9mmx2.5mm and up-sampled to 2.5mm isotropic.

Site 'C' was equipped with a 3T scanner with a 32-channel head coil. The scan acquisition parameters were same as that of site 'A', except for the number of slices (n=50) and GRAPPA=2 (instead of SENSE).

All *in vivo* acquisitions were pre-processed with standard procedures eddy, topup, b0 normalization and then registered pairwise per subject [17-19]. T1s were registered and transformed to the diffusion space. Brain extraction tool was used for skull stripping [19]. FAST white matter (WM) segmentation was performed using the T1 for the in vivo data [20]. Note that there were three pairs of pre-processed acquisitions in total.

The pair of data from Subject 1 along with the histology data set was used for the training of NSDN. The pairs of data from Subject 2 and 3 were used for quantitative and qualitative evaluation of the network. No site 'C' data were used in training.

## 3   Method: Network Design

Our proposed null space architecture is motivated by the linear algebra null spaces in that we need to design/constrain the aspect of the network that has no impact on the outcome. This work is inspired by [21] in which a person re-identification classification problem in computer vision was addressed using a Siamese architecture deep network. The novelty of our approach is that we use paired (but unlabeled) data to train the data-driven network to ignore potential features that would lead it to differentiate between the paired data.

The proposed network design takes three inputs of 8$^{th}$ order spherical harmonic (SH) coefficients (Fig 2). Each input provides an orthonormal representation of the DW-MRI signal and is known to characterize the angular diffusivity signal well [22]. The network outputs a 10<sup>th</sup> order SH FOD. The base network consists of five fully connected layers; the numbers of neurons per layer are 45, 400, 66, 200 and 66 in the respective order. Activation functions of 'ReLU' have only been used for the first two layers. They have not been used for the remainder of the layers to allow for negativity in the network because SH coefficients need not be positive. The three outputs obtained by the network are merged with a common loss function which optimizes on the assumption that the pairwise difference should be zero given the subject is the same and there should not be a difference in the FOD being predicted. For implementation simplicity, a modified weighted square loss function was defined as (here in $\lambda = 1$):

$$L = \frac{1}{m}\sum_{i=1}^{m}(y_{true_i} - y_{pred_i})^2 + \lambda (P_{a_i} - P_{b_i})^2 , \qquad (1)$$

where m is the total number of samples. $P_a$ and $P_b$ are paired *in vivo* voxels.

A sample size of 37,648 pairs of paired WM voxels were extracted from subject 1 using the acquisitions from site 'A' and site 'B'. A random selection of 37,648 data points was made from the training data set of the histology voxels. While training the network a K-fold cross-validation was used with K=5. The cross-validation set size was set to 0.2. 'RMSProp' was used as the optimizer of the network [23]. The number of iterations was determined at 3 using cross-validation. A batch size of 100 has been used.

To evaluate the performance, we use Angular Correlation Coefficient (ACC) which describes the correlation between two FOD's on a scale of -1 to 1 [6], where '1' is the best outcome.

## 4  Results

The median of the ACC computed from the blind set of 7,272 augmented histology voxels for CSD, DN and NSDN were 0.7965, 0.8165, and 0.8281, respectively. Non-parametric signed rank test for all pairs of distributions were found to be p < 0.01. Qualitatively, we explore the results relative to the truth voxel in Fig 3. At 25<sup>th</sup> percentile it can be observed that CSD and DN show a crossing fiber structure when compared to HFOD. NSDN is representative of more similar single fiber structure of histology. At 50<sup>th</sup> percentile CSD tends to show a crossing fiber structure, while DN and NSDN show a higher ACC and are like the structure of histology. At 75<sup>th</sup> percentile all three methods closely resemble the histology.

For subject 2, the histogram distribution of ACC for NSDN is most skewed (towards higher ACC) compared to DN and CSD (Fig 4A). The median values for the ACC distributions of CSD, DN and NSDN are: 0.67, 0.74 and 0.82. The gain in performance is (calculated by the difference of the medians) is 21.19% for (CSD, NSDN) and 10.09% for (DN, NSDN). Non-parametric signed rank test for all pairs of ACC metrics per voxels for subject 2 resulted in p< 0.001.

Qualitatively, we explore the spatial diffusion inferred structure in the WM of the frontal lobe of the middle axial brain slice (Fig 5). CSD (A & D) show low correlation and spurious fibers in the crossing fiber regions. DN (B & E) improves correlation in crossing fiber regions however NSDN (C & F) shows the highest correlation for crossing fiber regions. For single fibers, all three methods show high correlation.

In the quantitative results for subject 3, we observe that the skewed distribution towards higher ACC for NSDN is the highest as compared to both the other methods (Fig 4B). The median for three distributions of CSD, DN, and NSDN are 0.62, 0.67 and 0.74. The performance gain of NSDN over CSD is 16.08% and DN is 10.41%. Non-parametric signed rank test for all pairs of voxels for subject 3 show $p < 0.001$.

## 5   Conclusion

The NSDN method for reconstructing local fiber architecture is (1) more accurate when compared to histologically defined FODs, (2) more reproducible qualitatively and quantitatively on scan-rescan data, and (3) more reproducible on previously unseen scanners. While histological-MRI paired datasets are exceedingly rare, scan-rescan data are ubiquitous and often acquired as part of multi-site studies. The NSDN method provides a natural framework for harmonization that can use already acquired scan-rescan data to ensure that analysis methods are as reproducible across all sites. A much wider comparative study with multiple different HARDI methods and using multiple scanners is warranted. It would be interesting to explore the impact of including data from diffusion phantoms to enhance the diversity of signals captured in a data-driven approach.

While this work focused on DW-MRI, the NSDN approach can naturally be applied to other deep learning-based networks with two relatively simple modifications. First, one needs to construct a multiple channel network graph of the same form as "base network," but with shared weights for all channels and without cross-connections between the channels. This will ensure that one input can be placed per channel and all inputs will see the "same" base network. Second, the loss function needs to be modified so it combines a traditional loss with a reproducibility loss. The traditional loss comes with (without loss of generality) from the first channel's output relative to a traditionally provided truth dataset. The reproducibility loss is then computed by a metric of reproducibility between the remaining channels (herein a weighted squared error metric, but Dice, surface distance, etc. could be used as appropriate for the datatype). The potential synergies with data augmentation and neighborhood information have yet to be explored.

### ACKNOWLEDGMENETS

This work was supported by R01EB017230 (Landman). This work was conducted in part using the resources of the Advanced Computing Center for Research and Education at Vanderbilt University, Nashville, TN. This project was supported in part by the National Center for Research Resources, Grant UL1 RR024975-01, and is now at the National Center for Advancing Translational Sciences, Grant 2 UL1 TR000445-06. The content is solely the responsibility of theauthors and does not necessarily represent the official views of the NIH. This work has been supported by Nvidia with supplement of hardware resources (GPU's). Glyph visualizations were supported using [26].

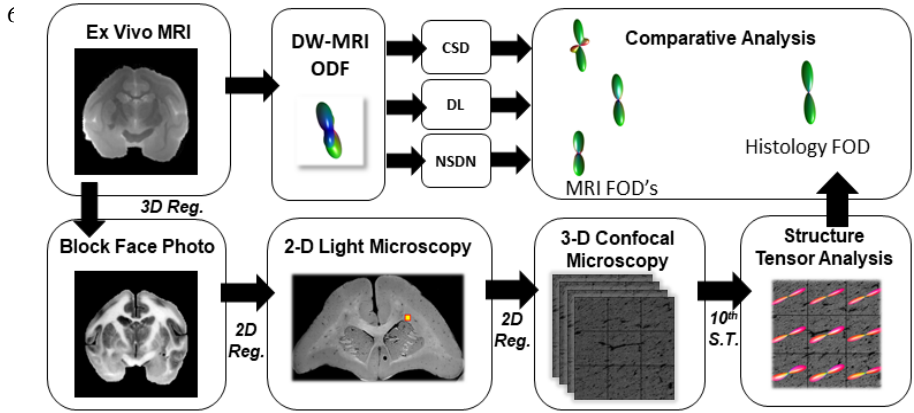

**Fig. 1.** Formation of the training dataset where 2D histology was performed on the squirrel monkey brains and FOD's were constructed per voxel basis using ensemble structure tensor analysis which correspond to *ex vivo* MRI acquisition of the squirrel monkey brains. Comparative analyses were performed between the reconstructed histology FOD's and FOD's from CSD, DN and NSDN.

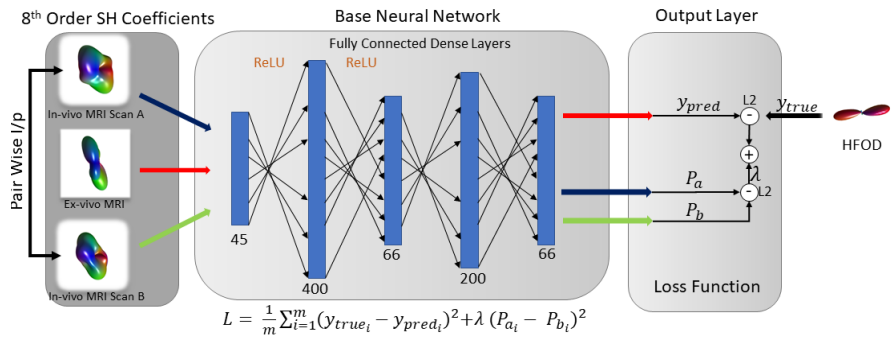

$$L = \frac{1}{m}\sum_{i=1}^{m}(y_{true_i} - y_{pred_i})^2 + \lambda \, (P_{a_i} - P_{b_i})^2$$

**Fig. 2.** Network design for the null space architecture. The architecture depicts how pairwise inputs of in-vivo voxels can be incorporated in a deep neural net architecture and can be added in the loss function as a noise enhancement/augmentation technique.

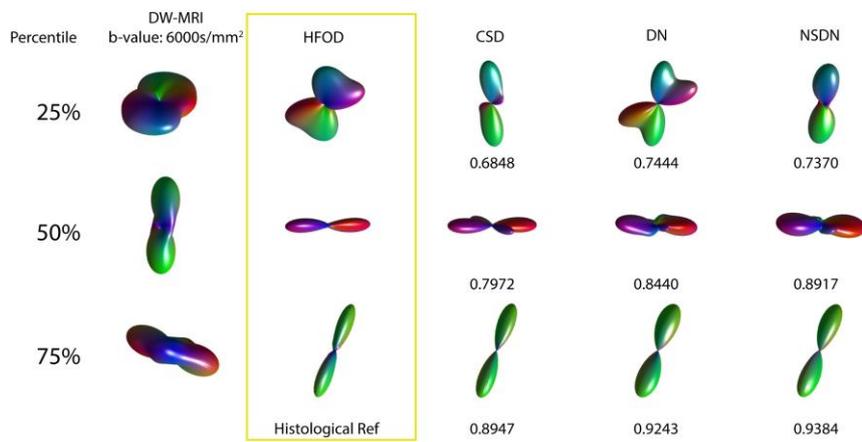

**Fig. 3.** Representative voxels are shown for the 25th, 50th, and 75th percentiles of CSD ACC along with corresponding DW-MRI, DN and NSDN glyphs and ACC's.

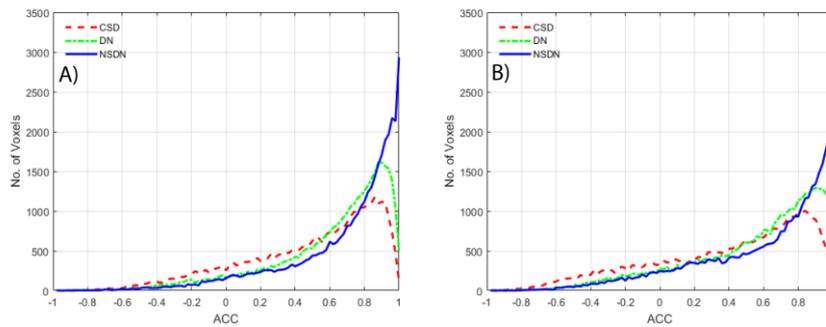

**Fig. 4.** A.) Histogram peaks of ACC per bin distributed over 100 bins for subject 2 of CSD, DN and NSDN. B) Histogram peak of ACC per bin distributed over 100 bins for subject 3 of CSD, DN and NSDN.

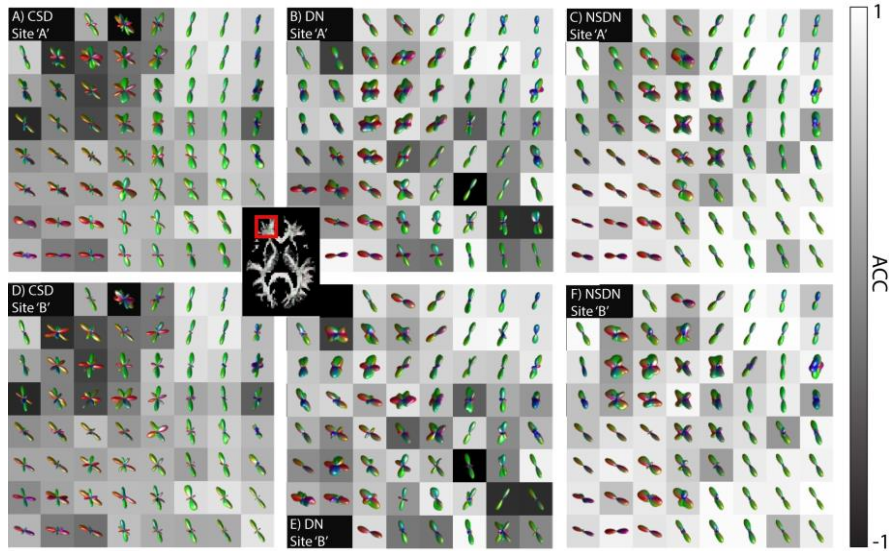

**Fig. 5.** The selected ROI shows left side frontal lobe of WM. The image underlay for the ROI's shown is Angular Correlation Coefficient (ACC) to indicate areas of agreement between scan-rescan (vertical pairs). CSD shows high correlation for core white matter where single fiber orientation exists (observer diagonal pattern of high ACC). DN shows increased correlation over a broader region which encompasses crossing fibers. NSDN shows a higher correlation across the most extended anatomical area.

# 6 References

1. Le Bihan, D., *Looking into the functional architecture of the brain with diffusion MRI.* Nature Reviews Neuroscience, 2003. **4**(6): p. 469.
2. Basser, P.J., J. Mattiello, and D. LeBihan, *MR diffusion tensor spectroscopy and imaging.* Biophysical journal, 1994. **66**(1): p. 259-267.
3. Tournier, J., F. Calamante, and A. Connelly. *How many diffusion gradient directions are required for HARDI.* in *Proc. Intl. Soc. Mag. Reson. Med*. 2009.
4. Tournier, J.-D., F. Calamante, and A. Connelly, *Robust determination of the fibre orientation distribution in diffusion MRI: non-negativity constrained super-resolved spherical deconvolution.* Neuroimage, 2007. **35**(4): p. 1459-1472.
5. Tuch, D.S., *Q-ball imaging.* Magnetic resonance in medicine, 2004. **52**(6): p. 1358-1372.
6. Anderson, A.W., *Measurement of fiber orientation distributions using high angular resolution diffusion imaging.* Magnetic Resonance in Medicine, 2005. **54**(5): p. 1194-1206.

7. Jansons, K.M. and D.C. Alexander, *Persistent angular structure: new insights from diffusion magnetic resonance imaging data.* Inverse problems, 2003. **19**(5): p. 1031.
8. Gorczewski, K., S. Mang, and U. Klose, *Reproducibility and consistency of evaluation techniques for HARDI data.* Magnetic Resonance Materials in Physics, Biology and Medicine, 2009. **22**(1): p. 63.
9. Nath, V., et al. *Comparison of multi-fiber reproducibility of PAS-MRI and Q-ball with empirical multiple b-value HARDI.* in *Medical Imaging 2017: Image Processing*. 2017. International Society for Optics and Photonics.
10. Helmer, K., et al. *Multi-site study of diffusion metric variability: effects of site, vendor, field strength, and echo time on regions-of-interest and histogram-bin analyses.* in *Medical Imaging 2016: Biomedical Applications in Molecular, Structural, and Functional Imaging*. 2016. International Society for Optics and Photonics.
11. Huo, J., et al., *Between-scanner and between-visit variation in normal white matter apparent diffusion coefficient values in the setting of a multi-center clinical trial.* Clinical neuroradiology, 2016. **26**(4): p. 423-430.
12. Mirzaalian, H., et al. *Harmonizing diffusion MRI data across multiple sites and scanners.* in *International Conference on Medical Image Computing and Computer-Assisted Intervention*. 2015. Springer.
13. Mirzaalian, H., et al., *Inter-site and inter-scanner diffusion MRI data harmonization.* NeuroImage, 2016. **135**: p. 311-323.
14. Fortin, J.-P., et al., *Harmonization of multi-site diffusion tensor imaging data.* Neuroimage, 2017. **161**: p. 149-170.
15. Stolp, H., et al., *Voxel-wise comparisons of cellular microstructure and diffusion-MRI in mouse hippocampus using 3D Bridging of Optically-clear histology with Neuroimaging Data (3D-BOND).* Scientific reports, 2018. **8**(1): p. 4011.
16. Krizhevsky, A., I. Sutskever, and G.E. Hinton. *Imagenet classification with deep convolutional neural networks.* in *Advances in neural information processing systems*. 2012.
17. Andersson, J.L., S. Skare, and J. Ashburner, *How to correct susceptibility distortions in spin-echo echo-planar images: application to diffusion tensor imaging.* Neuroimage, 2003. **20**(2): p. 870-888.
18. Andersson, J.L. and S.N. Sotiropoulos, *An integrated approach to correction for off-resonance effects and subject movement in diffusion MR imaging.* Neuroimage, 2016. **125**: p. 1063-1078.
19. Smith, S.M., *Fast robust automated brain extraction.* Human brain mapping, 2002. **17**(3): p. 143-155.
20. Zhang, Y., M. Brady, and S. Smith, *Segmentation of brain MR images through a hidden Markov random field model and the expectation-maximization algorithm.* IEEE transactions on medical imaging, 2001. **20**(1): p. 45-57.
21. Li, S., et al. *A discriminative null space based deep learning approach for person re-identification.* in *Cloud Computing and Intelligence Systems (CCIS), 2016 4th International Conference on*. 2016. IEEE.

22. Descoteaux, M., et al., *Apparent diffusion coefficients from high angular resolution diffusion imaging: Estimation and applications.* Magnetic Resonance in Medicine, 2006. **56**(2): p. 395-410.
23. Hinton, G., N. Srivastava, and K. Swersky, *Neural Networks for Machine Learning-Lecture 6a-Overview of mini-batch gradient descent.* 2012, Coursera Lecture slides.
24. Schilling, Kurt, et al. "Comparison of 3D orientation distribution functions measured with confocal microscopy and diffusion MRI." Neuroimage 129 (2016): 185-197.
25. Nath, Vishwesh, et al. " Deep Learning Captures More Accurate Diffusion Fiber Orientations Distributions than Constrained Spherical Deconvolution" ISMRM 2018, Paris, France
26. Justin Blaber, Kurt Schilling, & Bennett Landman. (2018, March 6). justinblaber/dwmri_visualizer: First release of dwmri_visualizer (Version v1.0.0). Zenodo. http://doi.org/10.5281/zenodo.1191107